\documentclass[remotesensing,article,accept,moreauthors,pdftex]{Definitions/mdpi} 

\usepackage[ruled,vlined]{algorithm2e}
\usepackage{amsmath}
\usepackage{longtable}
\usepackage{makecell, boldline}
\usepackage{multirow}

\firstpage{1} 
\pubvolume{xx}
\issuenum{1}
\articlenumber{5}
\pubyear{2020}
\copyrightyear{2020}
\history{Received: date; Accepted: date; Published: date}
\updates{yes} % If there is an update available, un-comment this line

\Title{A Two-Stream Symmetric Network with Bidirectional Ensemble for Aerial Image Matching}

\Author{Jae-Hyun Park $^{1,\dagger}$\orcidA{}, Woo-Jeoung Nam $^{1,\dagger}$\orcidB{} and Seong-Whan Lee $^{1,2,}$*\orcidC{}}

\AuthorNames{Jae-Hyun Park, Woo-Jeoung Nam and Seong-Whan Lee}

\address{%
$^{1}$ \quad Department of Computer and Radio Communications Engineering, Korea University, Anam-dong, Seongbuk-gu, Seoul 02841,  Korea; jh$\_\_$park@korea.ac.kr (J.-H.P.); nwj0612@korea.ac.kr (W.-J.N.)\\
$^{2}$ \quad Department of Artificial Intelligence, Korea University, Anam-dong, Seongbuk-gu, Seoul 02841,  Korea}

\corres{Correspondence: sw.lee@korea.ac.kr; Tel.: +82-2-3290-3197}

\firstnote{These authors contributed equally to this work.} 

\abstract{In this paper, we propose a novel method to precisely match two aerial images that were obtained in different environments via a two-stream deep network. By internally augmenting the target image, the network considers the two-stream with the three input images and reflects the additional augmented pair in the training. As a result, the training process of the deep network is regularized and the network becomes robust for the variance of aerial images. Furthermore, we introduce an ensemble method that is based on the bidirectional network, which is motivated by the isomorphic nature of the geometric transformation. We obtain two global transformation parameters without any additional network or parameters, which alleviate asymmetric matching results and enable significant improvement in performance by fusing two outcomes. For the experiment, we adopt aerial images from Google Earth and the International Society for Photogrammetry and Remote Sensing (ISPRS). To quantitatively assess our result, we apply the probability of correct keypoints (PCK) metric, which measures the degree of matching. The qualitative and quantitative results show the sizable gap of performance compared to the conventional methods for matching the aerial images. All code and our trained model, as well as the dataset are available online.}


\begin{document}

%%%%%%%%% Introduction
\section{Introduction}
\unskip
    \subsection{Motivation}
\begin{figure}[H]
    \centering
			\includegraphics[width=15.5cm]{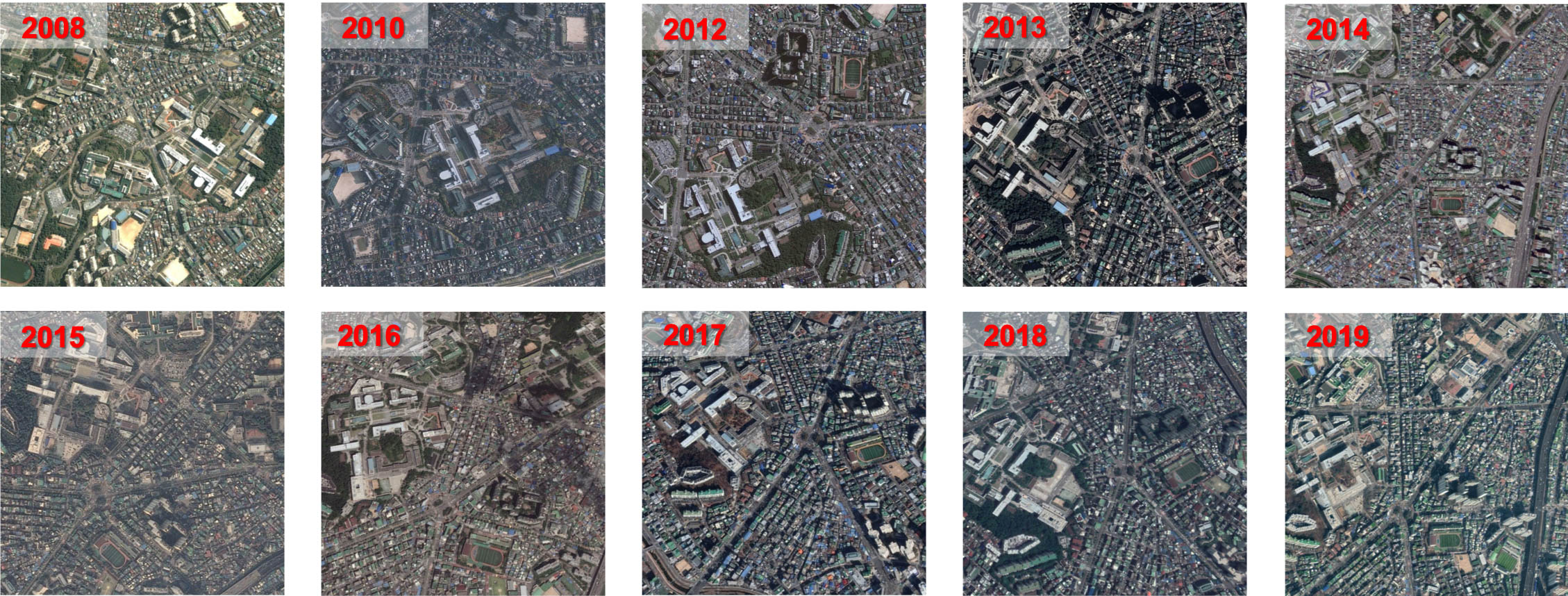}
			\caption{{Variance in the aerial image data.} We captured images that were obtained at different times, viewpoints and by different sensors. These images represent the same place but are visually different, which causes degradation in~performance.}
			\label{fig:fig_0}
	\end{figure}

    Aerial image matching is a geometric process of aligning a source image with a target image. Both images display the same scene but are obtained in different environments, such as time, viewpoints and sensors. It also a prerequisite of a variety of aerial image tasks such as change detection, image fusion, and~image stitching. Since it can have a significant impact on the performance of the following tasks, it is an extremely important task. As~shown in Figure~\ref{fig:fig_0}, various environments have considerable visual differences of land-coverage, weather, and~objects.
    The variance in the aerial images causes degradation of the matching precision. In~conventional computer vision approaches, correspondences between two images are computed by the hand-crafted algorithm (such as SIFT~\cite{Lowe04distinctiveimage}, SURF~\cite{Bay_surf:speeded}, HOG~\cite{Dalal05histogramsof}, and~ASIFT~\cite{Morel:2009:ANF:1658384.1658390}), followed by estimating the global geometric transformation using RANSAC~\cite{Fischler:1981:RSC:358669.358692} or Hough transform~\cite{Leibe_08_ijcv, Lamdan_88_cvpr}. However, these approaches are not very successful for aerial images due to their high-resolution, computational costs, large-scale transformation, and~variation in the~environments.
    
\begin{figure}[H]
		    \centering
			\includegraphics[width=12cm]{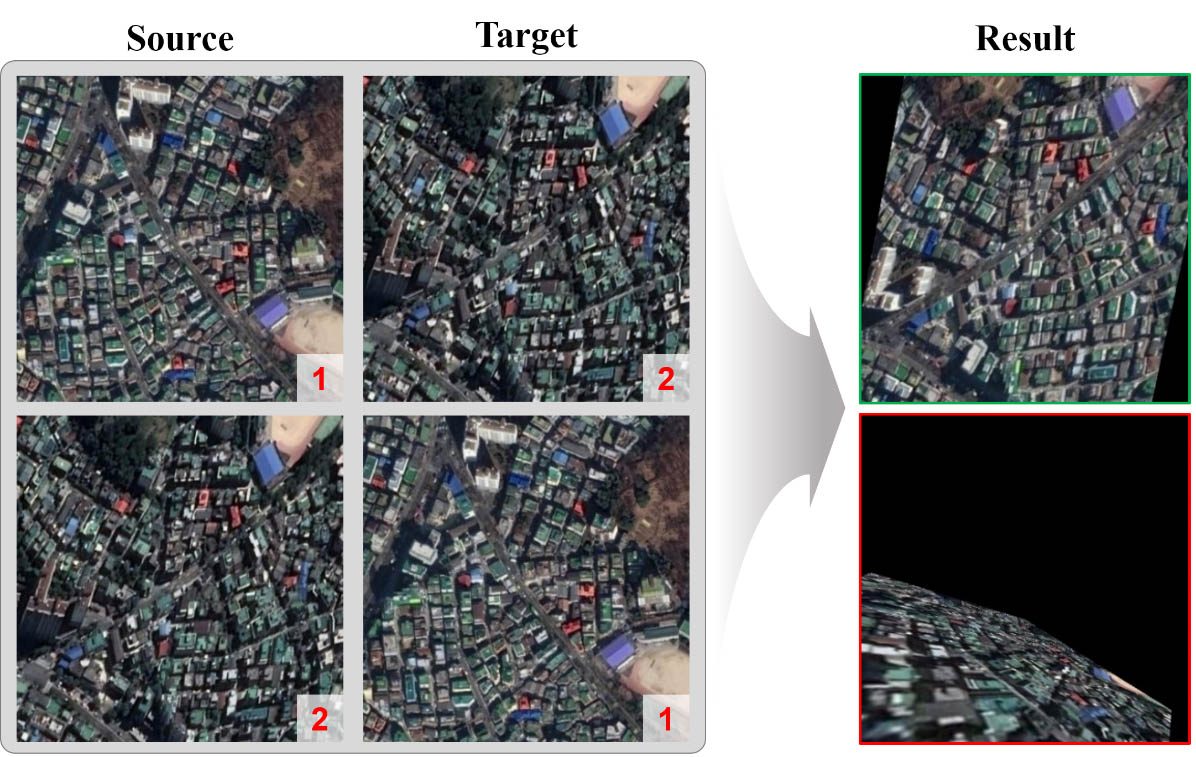}
			\caption{{Asymmetric matching result.} When image 1 and image 2 enter into source and target respectively, the~matching process is successful. In~the opposite case, however, it completely~fails.}
			\label{fig:asymm} 
	\end{figure}

    Another problem with aerial image matching is the asymmetric result. As~aforementioned, there are tons of aerial image matching methods~\cite{Lowe04distinctiveimage,Bay_surf:speeded,Dalal05histogramsof,Morel:2009:ANF:1658384.1658390,Fischler:1981:RSC:358669.358692,Leibe_08_ijcv, Lamdan_88_cvpr}. Notwithstanding, these methods~\cite{Lowe04distinctiveimage,Bay_surf:speeded,Dalal05histogramsof,Morel:2009:ANF:1658384.1658390,Fischler:1981:RSC:358669.358692,Leibe_08_ijcv, Lamdan_88_cvpr} have overlooked the consistency of matching flow. i.e.,~most methods consider only one direction of the matching flows (from source to target). It causes asymmetric matching results and degradation of the overall performance. In~Figure~\ref{fig:asymm}, it illustrates a failure case when the source image and the target image are~swapped.
    
   Many computer vision tasks have been applied and developed in real life~\cite{Dihua:2002,face_tracking, Roh:2010, Roh:2007, Nighttime, Kang:2014,SUK20103059,Jung:2004,Hwang:2000, Park:2007,Maeng:2011, PARK:2005, PARK:2004767,SUK:2011,SONG1996329,ROH:2000,book}. Because~deep n\scalebox{.94}[1.0]{eural networks (DNNs) have shown impressive performance in real-world computer vision tasks~\cite{Alex_12_NIPS, Girshick_15_FAST, Long_2015_CVPR, Goodfellow_14_NIPS}}, several approaches apply DNNs to overcome the limitation of traditional computer vision methods for matching the images. The~Siamese network~\cite{Koch_2015_SiameseNN, Chopra_05_learninga, Altwaijry_2016_CVPR, Melekhov2016SiameseNF} has been extensively applied to extract important features and to match image-patch pairs~\cite{Simo-Serra_2015_ICCV, Zagoruyko_2015_CVPR, Han_2015_CVPR}. Furthermore, several works~\cite{Rocco_2017_CVPR, Rocco_2018_CVPR, Seo_2018_ECCV} apply an end-to-end manner in the geometric matching area. While numerous matching tasks have been actively explored with deep learning, few approaches utilize DNNs in aerial image matching~areas.
    
\begin{figure}[H]
		    \centering
			\includegraphics[width=15.5cm]{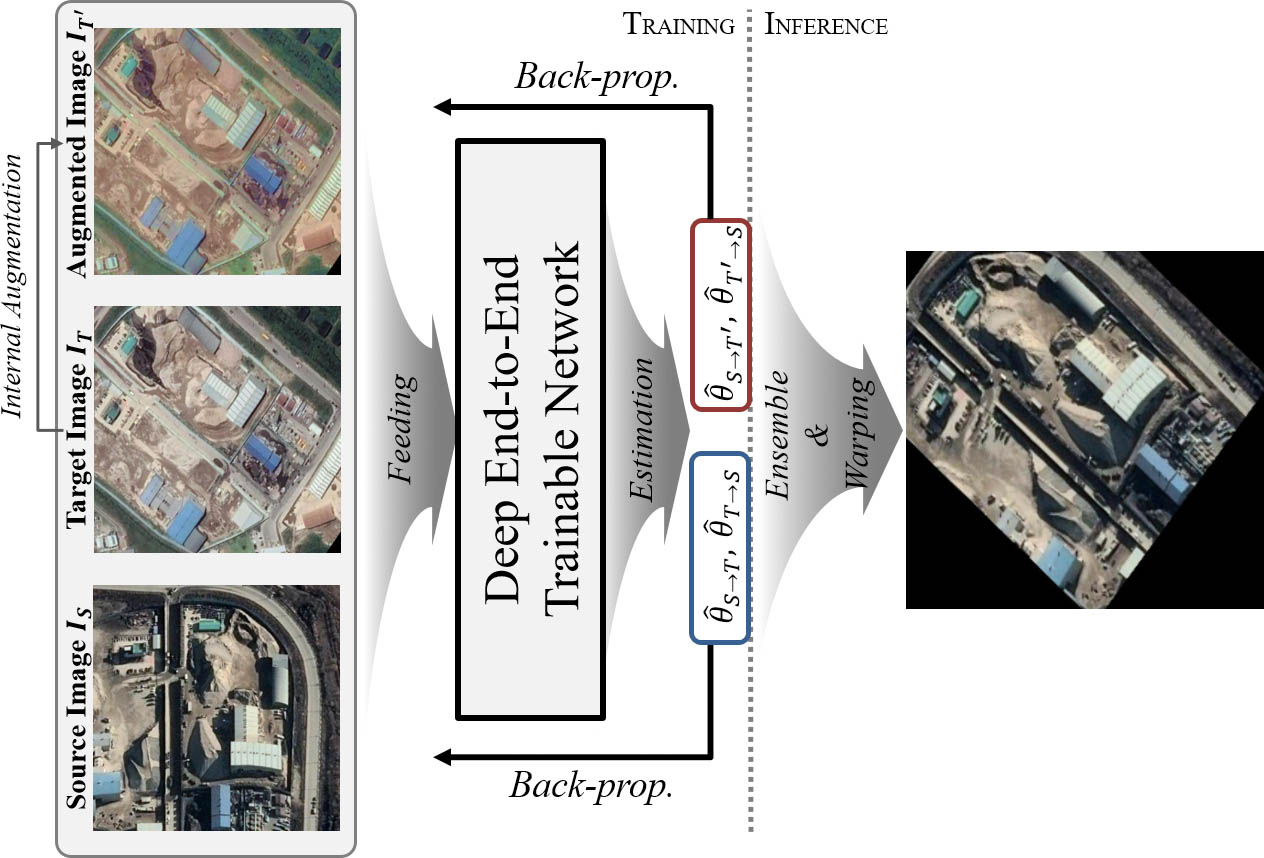}
			\caption{{Overview of the proposed network.} Our network directly estimates the outcomes $(\hat{\theta}_{S \rightarrow T}, \hat{\theta}_{T \rightarrow S}, \hat{\theta}_{S \rightarrow T'},\hat{\theta}_{T' \rightarrow S})$, where $\hat{\theta}_{S \rightarrow T}$ and $\hat{\theta}_{T \rightarrow S}$ are the global transformation parameters that transform $I_S$ to $I_T$, vice~versa, and~($\hat{\theta}_{S \rightarrow T'}$, $\hat{\theta}_{T' \rightarrow S}$) are those between $I_S$ and $I_{T'}$. Subsequently, the~outcomes are employed for the backpropagation in the training procedure. In~the inference procedure, we warp $I_S$ to $I_T$ using the final ensembled parameters. }
			\label{fig:fig_1}
	\end{figure}

    In this work, we utilize a deep end-to-end trainable matching network and design a two-stream architecture to address the variance in the aerial images obtained in diverse environments. By~internally augmenting the target image and considering the three inputs, we regularize the training process, which produces a more generalized deep network.~Furthermore, our method is designed as a bidirectional network with an efficient ensemble manner. Our ensemble method is inspired by the isomorphic nature of the geometric transformation. We apply this method in our inference procedure without any additional networks or parameters. The~ensemble approach also assists in alleviating the variance between estimated transformation parameters from both directions. Figure~\ref{fig:fig_1} illustrates an overview of our proposed~method.

	\subsection{Contibutions}
	To sum up, our contributions are~three-fold:
	\begin{itemize}[leftmargin=*,labelsep=5mm]

		\item For aerial image matching, we propose a deep end-to-end trainable network with a two-stream architecture. The~three inputs are constructed by internal augmentation of the target image, which regularizes the training process and overcomes the shortcomings of the aerial images due to various capturing~environments.
		
		\item We introduce a bidirectional training architecture and an ensemble method, inspired by the isomorphism of the geometric transformation. It alleviates the asymmetric result of image matching. The~proposed ensemble method assists the deep network to become robust for the variance between estimated transformation parameters from both directions and shows improved performance in evaluation without any additional network or~parameters.
		
		\item Our method shows more stable and precise matching results from the qualitative and quantitative assessment. In~the aerial image matching domain, we first apply probability of correct keypoints (PCK) metrics [44] to objectively assess quantitative performance with a large volume of aerial images. Our dataset, model and source code are available at \url{https://github.com/jaehyunnn/DeepAerialMatching}.
	\end{itemize}
	
	\subsection{Related~Works}
	\label{sec:related}
	In general, the~image matching problem has been addressed in two types of methods: area-based methods and feature-based methods~\cite{Brown:1992:SIR:146370.146374, Zitova03imageregistration}. The~former methods investigate the correspondence between two images using pixel intensities. However, these methods are vulnerable to noise and variation in illumination. The~latter methods extract the salient features from the images to solve these~drawbacks.
	
	Most classical pipelines for matching two images consist of three stages, (1) feature extraction, (2) feature matching, and (3) regression of transformation parameters. As~conventional matching methods, hand-crafted algorithms~\cite{Lowe04distinctiveimage,Bay_surf:speeded,Dalal05histogramsof,Morel:2009:ANF:1658384.1658390} are extensively used to extract local features. However, these methods often fail for large changes in~situations, which is attributed to the lack of generality for various tasks and image~domains.
	
\textls[-10]{Convolutional neural networks (CNNs) have shown tremendous strength for extracting high-level features to solve various computer vision tasks, such as semantic segmentation~\cite{Long_2015_CVPR,Chen_2018_ECCV}, object detection~\cite{Girshick_15_FAST, SNIPER_18}, classification~\cite{Alex_12_NIPS, Hu_2018_CVPR}, human action recognition~\cite{ P-S, Yang:2007}, and~matching. In~the field of matching, E. Simo-Serra et al.~\cite{Simo-Serra_2015_ICCV} learned local features based on image-patch with a Siamese network and use the L2-distance for the loss function. X. Han et al.~\cite{Han_2015_CVPR} proposed a feature network and metric network to match two image patches. S. Zagoruyko et al.~\cite{Zagoruyko_2015_CVPR} expanded the Siamese network in two-streams: surround stream and central stream.~K.-M. Yi et al.~\cite{Yi_2016_ECCV} proposed a framework that includes detection, orientation, estimation, and~description by mimicking SIFT~\cite{Lowe04distinctiveimage}. \mbox{H. Altwaijry et al.}~\cite{Altwaijry_2016_CVPR} performed ultra-wide baseline aerial image matching with a deep network and spatial transformer module~\cite{Max_2015_STN}. H. Altwaijry et al.~\cite{Altwaijry_2016_BMVC} also proposed a deep triplet architecture that learns to detect and match keypoints with 3-D keypoints ground-truth extracted by VisualSFM~\cite{Wu_2011_CVPR, Wu_2013_SFM}.  \mbox{I. Rocco et al.}~\cite{Rocco_2017_CVPR} first proposed a deep network architecture for geometric matching, and~demonstrated the advantage of a deep end-to-end network by achieving 57\% PCK score in the semantic alignment. This method constructs a dense-correspondence map using two image features and directly regress the transformation parameters. These researchers further proposed a weakly-supervision approach that does not require any additional ground-truth for training~\cite{Rocco_2018_CVPR}. P. Seo et al.~\cite{Seo_2018_ECCV} applied an attention mechanism with an offset-aware correlation (OAC) kernel based on~\cite{Rocco_2017_CVPR} and achieved a 68\% PCK~score.}
	
	Although these works show meaningful results, their accuracy or computational costs for aerial image matching require improvement. Therefore, we compose a matching network that is suitable for aerial images by pruning the factors that degrade~performance.

%%%%%%%%% Method
\section{Materials and~Methods}
\label{sec:proposed}
\begin{figure}[H]
		\centering
			\includegraphics[width=15.5cm]{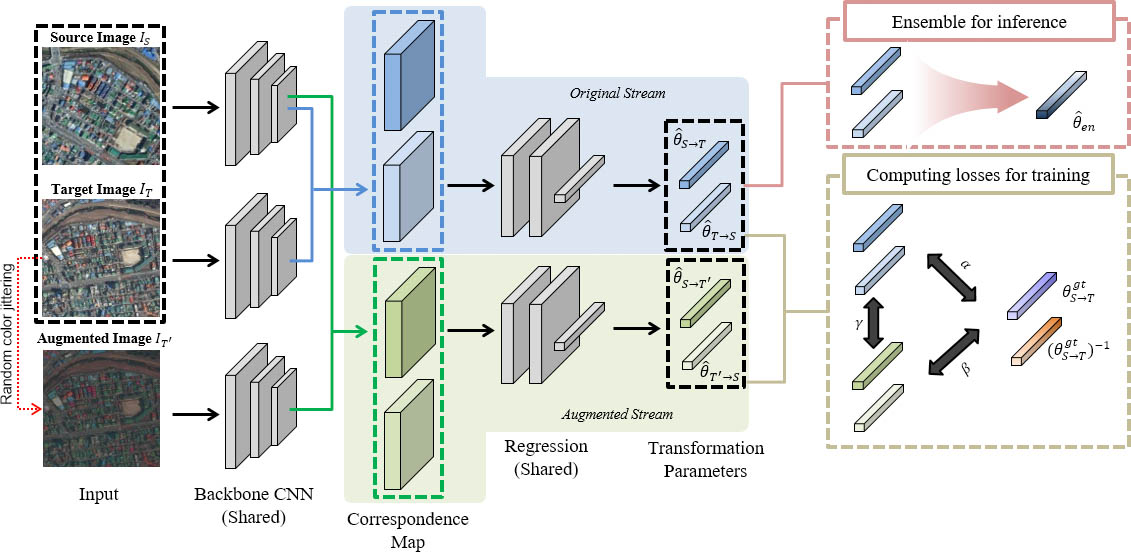}
			\caption{\textls[-10]{{Overall architecture of the proposed network.} Architecture has four stages: internal augmentation, feature extraction, matching, and~regression. First, the~target image is augmented using random color-jittering. Subsequently, the~source,} target, and~augmented images are passed through the backbone networks which share the weights, followed by the matching operations, which produces the correspondence maps. The~regression networks which also share the weights simultaneously output the geometric transformation parameters of the original pair $(I_S, I_T)$ and the augmented pair $(I_S, I_{T^{'}})$. We fuse the transformation parameters $(\hat{\theta}_{S \rightarrow T}, \hat{\theta}_{T \rightarrow S})$ for inference or compute the losses with the balance parameters $\alpha, \beta$, and~$\gamma$ for~training.}
			\label{fig:fig_2}
	\end{figure}

    We propose a deep end-to-end trainable network with a two-stream architecture and bidirectional ensemble method for aerial image matching. Our proposed network focuses on addressing the variance in the aerial images and asymmetric matching results. The~steps for predicting transformation are listed as follows: (1) internal augmentation, (2) feature extraction with the backbone network, (3) correspondence matching, (4) regression of transformation parameters, and~(5) application of ensemble to the multiple outcomes. In~Figure~\ref{fig:fig_2}, we present the overall architecture of the proposed~network.
    
	\subsection{Internal Augmentation for~Regularization}
\begin{figure}[H]
	    \centering
	    \includegraphics[width=15.5cm]{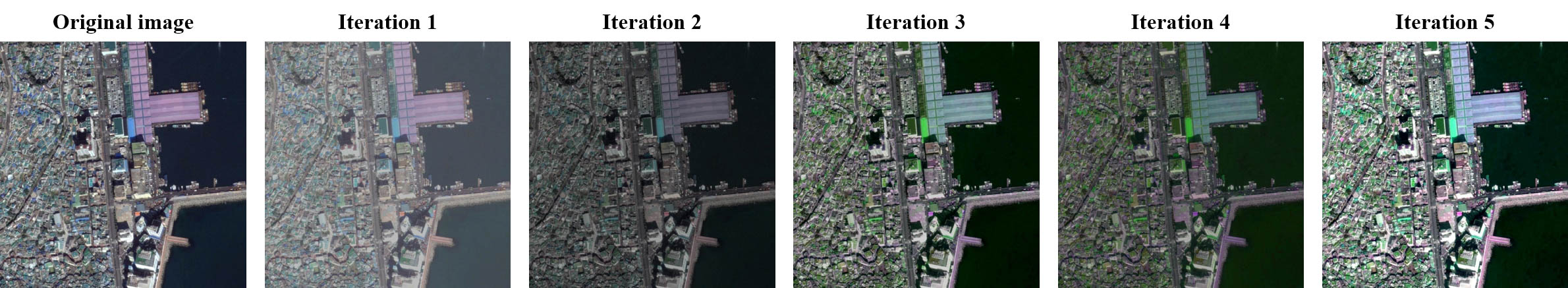}
	    \caption{{Internal augmented samples.} In every iteration of training, the~target image is augmented using random color-jittering. Therefore, in~every iteration, the~network considers a different augmented training~pair.}
	    \label{fig:augmenting}
	\end{figure}

    	The network considers two aerial images (source image $I_S$ and target image $I_T$) with different temporal and geometric properties as the input. By~using this original pair $(I_S, I_T)$ in the training process, the~deep network is trained by considering the relation of only two images obtained in different environments. However, this approach is insufficient for addressing the variance in the aerial images. Collecting various pair sets to solve these problems is expensive. To~address this issue, we augment the target image by internally jittering image color during the training procedure. The~network can be trained with various image pairs since the color of the target image is randomly jittered in every training iteration as shown in Figure~\ref{fig:augmenting}. This step has a regularization effect of the training process, which produces a more generally trained network. The~constructed three inputs are passed through a deep network. Subsequently, the~network directly and simultaneously estimates global geometrical transformation parameters for the original pair and augmented pair. Note that the~internal augmentation is only performed in the training procedure. In~inference procedure, we utilize a single-stream architecture without the internal augmentation process for computational~efficiency. 

	\subsection{Feature Extraction with Backbone~Network}\label{ssec:feature_extraction}
    	Given the input images $(I_S, I_T, I_{T'})\in \mathbb{R}^{h\times w\times d}$, we extract their feature maps $(f_S, f_T, f_{T'})\in \mathbb{R} ^{h'\times w'\times d'}$ by passing a fully-convolutional backbone network $\mathcal{F}$, which is expressed as follows:
\begin{equation}
    	\label{eq:eq_1}
    	\mathcal{F}: \mathbb{R} ^{h\times w\times d}\rightarrow\mathbb{R} ^{h'\times w'\times d'},
    	\end{equation}
    	where $(h, w, d)$ denote the heights, widths, and~dimensions of the input images and $(h', w', d')$ are those of the extracted features, respectively.
	
\begin{figure}[H]
    	    \centering
    	    \includegraphics[width=15.5cm]{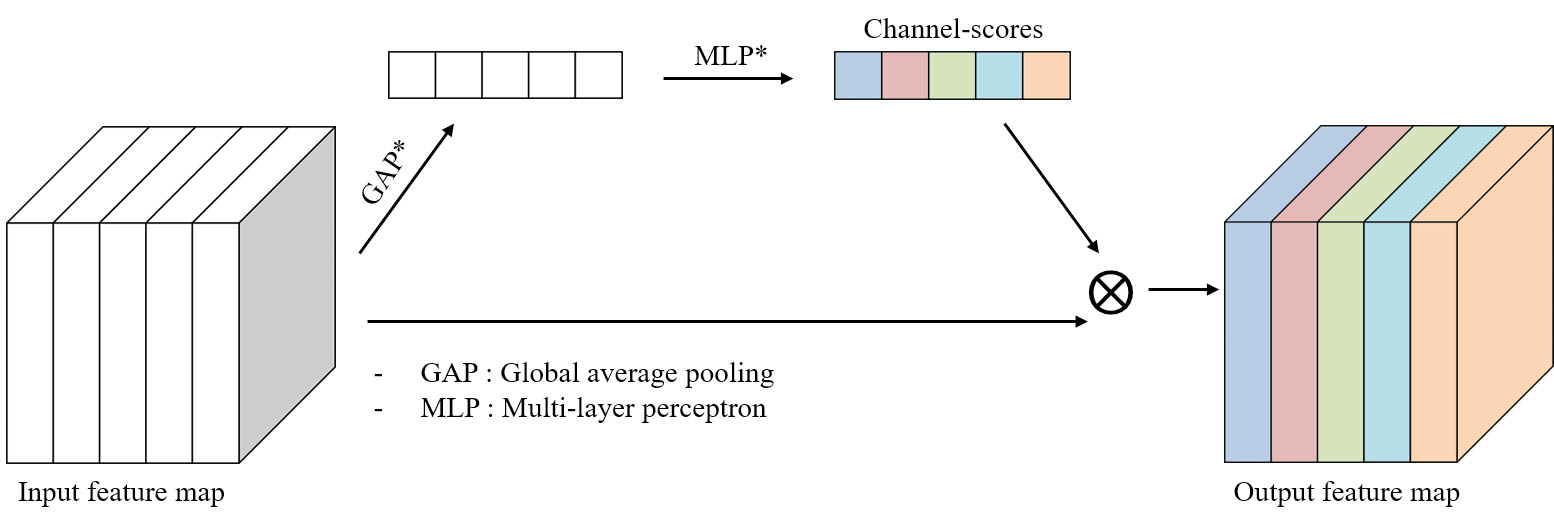}
    	    \caption{{Squeeze-and-Excitation (SE) block.} The input feature map is applied by global average pooling (GAP), followed by a multi-layer perceptron (MLP). The~input feature map is elementwise multiplied by the~channel-scores.}
    	    \label{fig:se_block}
    	\end{figure}

    	We investigate various models of the backbone networks, as~shown in Section~\ref{sec:experiments}. SE-ResNeXt101~\cite{Hu_2018_CVPR} add the Squeeze-and-Excitation (SE) block as the channel-attention module to ResNeXt101~\cite{Xie_2017_CVPR}, which has shown its superiority in~\cite{ILSVRC15}. Figure~\ref{fig:se_block} shows the SE-block. Therefore, we leverage SE-ResNeXt101 as the backbone network and empirically show that it has an important role in improving performance compared with other backbone networks. We utilize the image features extracted from layer-3 in the backbone network and apply L2-normalization to extracted~features.
    	
\unskip
    	
	\subsection{Correspondence~Matching}\label{ssec:matching}
    	As a method for computing a dense-correspondence map between two feature maps~\cite{Rocco_2017_CVPR}, the~matching function $\mathcal{C}$ is expressed as follows:
\begin{align}
    		\label{eq:eq_2}
    		c_{S\rightarrow T}(i, j, k)&=\mathcal{C}(f_S(i_k, j_k), f_T(i,j))\nonumber\\
    		& = f_T(i,j)^{f}_{S}(i_k, j_k),
    	\end{align}
    	where $c_{S\rightarrow T}$ is the dense-correspondence map that matches the source feature map $f_S$ to the target feature map $f_T$. $(i,j)$ and $(i_k,j_k)$ indicate the coordinate of each feature point in the feature maps. Each element in $c_{S\rightarrow T}$ refers to the similarity score between two~points.
    	
    	We construct the dense-correspondence map of the original pair and augmented pair. To~consider only positive values for ease of training, the~negative scores in the dense-correspondence map are removed by ReLU non-linearity, followed by~L2-normalization.
	
	\subsection{Regression of Transformation~Parameters}\label{ssec:regression}
    	The regression step is for predicting the transformation parameters. When the dense-correspondence maps are passed through the regression network $\mathcal{R}$, the~network $\mathcal{R}$ directly estimates the geometric transformation parameters as follows:
\begin{equation}
    	    \mathcal{R}: \mathbb{R} ^{h'\times w'\times (h'\times w')}\rightarrow\mathbb{R}^{DoF},
    	\end{equation}
    	where $(h', w')$ indicate the heights and widths of the feature maps, and~$DoF$ means the degrees of freedom of the transformation~model.
    	
        We adopt the affine transformation which has 6-$DoF$ and the ability to preserve straight lines. In~the semantic alignment domain~\cite{Rocco_2017_CVPR, Rocco_2018_CVPR, Seo_2018_ECCV}, thin-plate spline (TPS) transformation~\cite{TPS_89} which has 18-$DoF$ is used to improve the performance. However, it is not suitable in the aerial image matching domain, because~it produces large distortions of the straight lines (such as roads and boundaries of the buildings). Therefore, we infer the six parameters that handle the affine~transformation.
    	
	\subsection{Ensemble Based on Bidirectional~Network}\label{ssec:bi_direction}
    	The affine transformation is invertible due to its isomorphic nature. We take advantage of this characteristic to design a bidirectional network and apply an ensemble approach. Applying the ensemble method enables alleviating the variance in the aerial images and improvement in the matching performance without any additional networks or models.
	
    	\subsubsection{Bidirectional Network}
    	
		Inspired by its isomorphic nature, we expand the base architecture by adding a branch that symmetrically estimates the transformation in the opposite direction symmetrically. The~network yields the transformation parameters in both directions of each pair, i.e.,~$(\hat{\theta}_{S \rightarrow T}, \hat{\theta}_{T \rightarrow S})$ and $(\hat{\theta}_{S \rightarrow T'}, \hat{\theta}_{T' \rightarrow S})$. To~infer the parameters of another branch, we compute the dense-correspondence map in the opposite direction by using the same method as in Section \ref{ssec:matching}. 
        All dense-correspondence maps are passed through the identical regression network $\mathcal{R}$. Since we utilize a regression network for all cases, no additional parameters are needed in this procedure. The~proposed bidirectional network only adds a small amount of computational overhead compared with the base architecture.
      	\subsubsection{Ensemble} 
\begin{figure}[H]
			\centering
			\includegraphics[width=13cm]{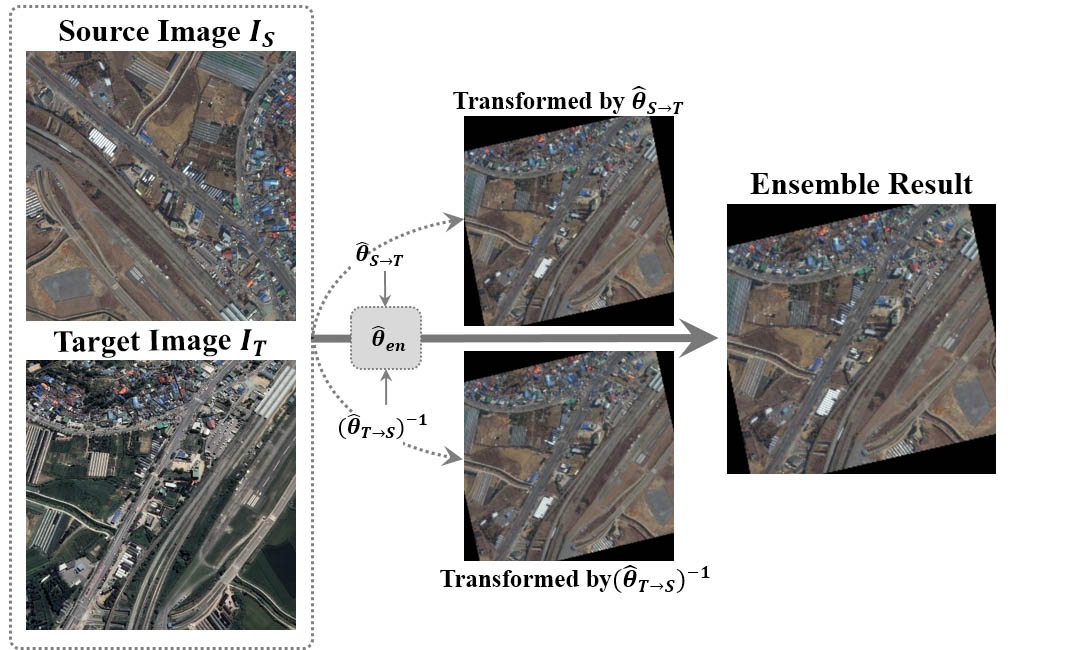}
			\caption{{Ensemble process of affine parameters.} The outcomes that correspond to the original pair are the transformation parameters $(\hat{\theta}_{S\rightarrow T}, \hat{\theta}_{T\rightarrow S})$ in two possible directions. Since the affine transformation is isomorphic, we can use the inverse of $\hat{\theta}_{T\rightarrow S}$ to warp the source image to the target image. Therefore, the~final transformation parameters are obtained by fusing these~parameters.}
			\label{fig:fig_3}
		\end{figure}

		In general, the~ensemble technique requires several additional different architectures and consumes additional time costs to train models differently. We introduce an efficient ensemble method without any additional architectures or models by utilizing the isomorphism of the affine transformation. 
	 Figure~\ref{fig:fig_3} illustrates the overview of the ensemble procedure. 
        $(\hat{\theta}_{T \rightarrow S})^{-1}$, which is the inverse of $\hat{\theta}_{T \rightarrow S}$, can be expressed as another transformation parameters in the direction from $I_S$ to $I_T$.
		To compute $(\hat{\theta}_{T \rightarrow S})^{-1}$, we convert $\hat{\theta}_{T \rightarrow S}$ into the homogeneous form:
\begin{equation}
		\label{eq:eq_4}
    		[a_1, a_2, t_x, a_3, a_4, t_y]\Longrightarrow 
    		\begin{bmatrix}
    		a_1 & a_2 & t_x\\
    		a_3 & a_4 & t_y\\
    		0 & 0 & 1
    		\end{bmatrix}.
		\end{equation}

		In the affine transformation parameters $[a_1, a_2, t_x, a_3, a_4, t_y]$, $a_1\sim a_4$ represent the scale, rotated angle and tilted angle, and~$(t_x, t_y)$ denotes the $(x$-axis, $y$-axis$)$ translation. We compute $(\hat{\theta}_{T \rightarrow S})^{-1}$ by converting the homogeneous form, as~shown in Equation~(\ref{eq:eq_4}). This inverse matrix denotes another affine transformation from $I_S$ to $I_T$. 
		As a result, we fuse the two sets of affine transformation parameters as follows:
\begin{equation}
		    \hat{\theta}_{en}=\mu (\hat{\theta}_{S \rightarrow T}, (\hat{\theta}_{T \rightarrow S})^{-1}),
		    \label{eq:ensemble}
		\end{equation}
		where $\mu (*)$ denotes the mean function for fusing two parameters. In~the various experiments, we apply three types of mean: arithmetic mean, harmonic mean and geometric mean. Empirically, arithmetic mean shows the best performance. In~the inference process, $\hat{\theta}_{en}$ warps the source image into the target image. Note that we fuse only parameters that correspond to the original pair since we use the original two-stream network in the inference procedure and do not utilize the ensembled parameters in the training procedure to maximize the ensemble~effects.
    	
\unskip
    	
	\subsection{Loss~Function}
		In the training procedure, we adopt the transformed grid loss~\cite{Rocco_2017_CVPR} as the baseline loss function. Given the predicted transformation $\hat{\theta}$ and the ground-truth $\theta^{gt}$, the~baseline loss function $l(\hat{\theta},\theta^{gt})$ is obtained by the following:
\begin{equation}
		l(\hat{\theta},\theta^{gt})=\frac{1}{N}\sum_{i,j=1}^Nd(\mathcal{T}_{\hat{\theta}}(x_i, y_j),\mathcal{T}_{\theta^{gt}}(x_i, y_j))^2,
		\end{equation}
		where $N$ is the number of grid points, $\mathcal{T}_{\hat{\theta}}(*)$ and $\mathcal{T}_{\theta^{gt}}(*)$ are the transforming operations parameterized by $\hat{\theta}$ and $\theta^{gt}$, respectively. To~achieve bidirectional learning, we add a term for training the additional branch to the baseline loss function. Formally, we define the proposed bidirectional loss of the original pair, $\mathcal{L}_{org}$, as~follows:
\begin{align}
        	\label{eq:eq_7}
        	\mathcal{L}_{org}=
        	&l(\hat{\theta}_{S \rightarrow T},\theta_{S \rightarrow T}^{gt}) + l(\hat{\theta}_{T \rightarrow S},(\theta_{S \rightarrow T}^{gt})^{-1}).
	    \end{align}

		Note that additional ground-truth information for the opposite direction is not required due to the isomorphism of the affine transformation. For~regularization of training, we add two terms utilizing the augmented pair:
\begin{align}
        	\label{eq:eq_8}
        	\mathcal{L}_{aug}=
        	&l(\hat{\theta}_{S \rightarrow T'},\theta_{S \rightarrow T}^{gt}) + l(\hat{\theta}_{T' \rightarrow S},(\theta_{S \rightarrow T}^{gt})^{-1}),
	    \end{align}
\begin{align}
        	\label{eq:eq_9}
        	\mathcal{L}_{id} =
        	&l(\hat{\theta}_{S \rightarrow T},\theta_{S \rightarrow T'}) +l(\hat{\theta}_{T \rightarrow S},\theta_{T' \rightarrow S}).
	    \end{align}

	    The augmented pair also share the ground-truth since the geometric relation between two images is equivalent to the original pair. The~identity term in Equation~(\ref{eq:eq_9}) induces training to ensure that the prediction values from the original pair and the augmented pair are equal. Our proposed final loss function is defined by the following:
	    \vspace{12pt}
\begin{align}
        	\label{eq:eq_10}
        	\mathcal{L} = 
        	&\alpha \cdot \mathcal{L}_{org} + 
        	\beta \cdot \mathcal{L}_{aug} + 
        	\gamma \cdot \mathcal{L}_{id},
	    \end{align}
	    where $(\alpha, \beta, \gamma)$ are the balance parameters of each loss term. In~our experiment, we set these parameters to (0.5, 0.3, 0.2), respectively. 

%%%%%%%%% Experiments
\section{Results}
	\label{sec:experiments}
	In this section, we present the implementation details, experiment settings, and~results. For~the quantitative evaluation, we compare the proposed method with other methods for aerial image matching. We further experiment with various backbone networks to obtain more suitable features for our work. We show the contributions of each proposed component in the ablation study section and the qualitative results of the proposed network compared with other~networks.
	
	\subsection{Implementation~Details}
	We implemented the proposed network using PyTorch~\cite{paszke2017automatic} and trained our model with the ADAM optimizer~\cite{kingma_2015_ICLR}, using a learning rate $5\times10^{-4}$ and a batch size of 10. We further performed data augmentation by generating the random affine transformation as the ground-truth. All input images were resized to $240\times240$.
    
    \begin{figure}[H]
		\centering
		\includegraphics[width=13cm]{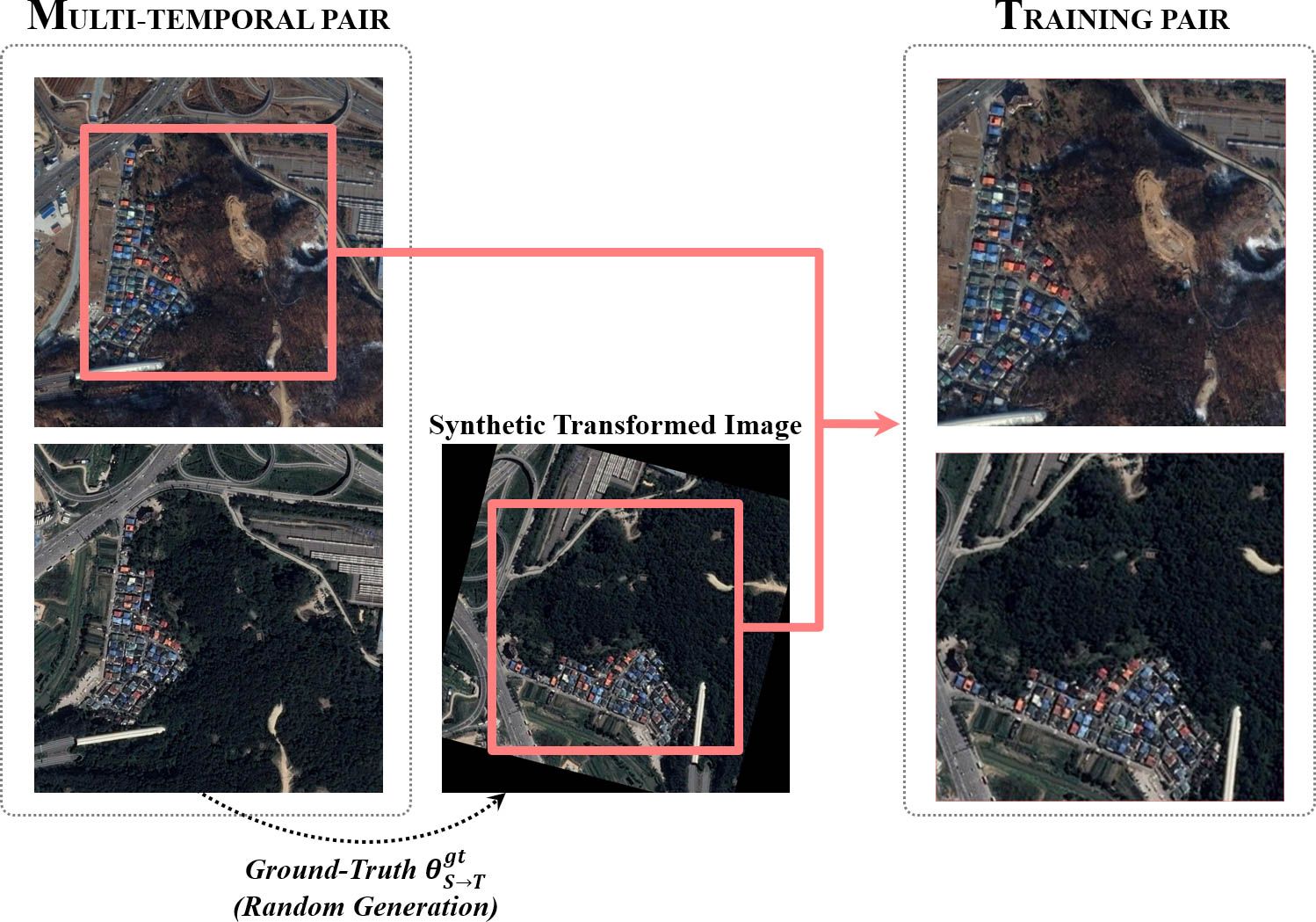}
		\caption{{Process of generating the training pairs.} In the training procedure, given a multi-temporal aerial image pair, we perform the transformation on the second image using the ground-truth $\theta_{S\rightarrow T}^{gt}$ which is randomly~generated.}
		\label{fig:fig_4}
	\end{figure}
    
	\subsection{Experimental~Settings}
	\subsubsection{Training}
    We generated the training input pairs by applying random affine transformations to the multi-temporal aerial image pairs captured in Google Earth. Since no datasets were annotated with completely correct transformation parameters between two images, we built the training dataset, 9000 multi-temporal aerial image pairs, and corresponding ground-truths. Basically, multi-tempral image pairs consisted of the image pairs which were taken at different times (2019, 2017, and~2015)  and by different sensors (Landsat-7, Landsat-8, WorldView, and~QuickBird). The~process of annotating ground-truth is as follows: (1) we employed the multi-temporal image pairs $(I,I')$ with the same region and viewpoint. (2) The first images in the multi-temporal aerial image pairs were center-cropped. (3) The second images are transformed by the randomly generated affine transformation $\theta^{gt}_{S \rightarrow T}$ which was used as a ground-truth and subsequently center-cropped. (4) The center-crop process was performed to exclude the black area that serves as noise after transformation. Figure~\ref{fig:fig_4} illustrates the process of generating training pairs and ground-truths. 
    In Algorithm \ref{alg:alg_1}, the~training procedure is detailed. It has $O(N)$ complexity with respect to the number of training pairs $N$. We train our model for 2-days on a single NVIDIA Titan V GPU. \vspace{12pt}

	% Training procedure
		\begin{algorithm}[H]
	     \label{alg:alg_1}
	           \SetAlgoLined
        \SetKwInOut{Input}{Input}
        \Input{Training aerial image dataset \textbf{\textit{D}} \\ Randomly initialized model $\mathcal{M}_w$}
        \SetKwInOut{Output}{Output}
        \Output{Trained model $\mathcal{M}_w$}
                 \For{epochs}
         {
                  \For{$(I, I')$ in \textbf{D}}
                  {
                 \texttt{{\#} Construct three inputs}\\
            $\theta^{gt}_{S \rightarrow T}=$ randomly generated transformation\;
            $I_S=$ center-cropped image of $I$\;
            $I_T=$ center-cropped image of $\mathcal{T}_{\theta^{gt}_{{S \rightarrow T}}}(I')$\;
            $I_{T^{'}}=$ color-jittered image of $I_T$\;
                       \texttt{{\#} Feed-forward}\\
            $\hat{\theta}_{S \rightarrow T}, \hat{\theta}_{T \rightarrow S}, \hat{\theta}_{S \rightarrow T^{'}}, \hat{\theta}_{T^{'} \rightarrow S}=\mathcal{M}_w(I_S,I_T,I_{T^{'}})$\;
                       \texttt{{\#} Compute loss}\\
            $L=\mathcal{L}(\hat{\theta}_{S \rightarrow T}, \hat{\theta}_{T \rightarrow S}, \hat{\theta}_{S \rightarrow T^{'}}, \hat{\theta}_{T^{'} \rightarrow S}, \theta^{gt}_{S \rightarrow T})$\;
                        \texttt{{\#} Backpropagation and update weights}\\
            $w = w-\eta(\frac{\partial L}{\partial w})$ \;
        }
         }
         \caption{Training~procedure.}
             \end{algorithm}

	\subsubsection{Evaluation}
	
	 To demonstrate the superiority of our method quantitatively, we evaluated our model using the PCK \cite{pck}, which was extensively applied in the other matching tasks~\cite{Ham_2016_CVPR, Han_2017_ICCV, Rocco_2017_CVPR, Rocco_2018_CVPR, Seo_2018_ECCV, Kim_2017_ICCV, Kim_2017_CVPR}. PCK metric is defined as follows:
\begin{equation}
		\label{eq:eq_10}
		PCK=\frac{\sum_{i=1}^n\sum_{p_i}1[d(\mathcal{T}_{\hat{\theta}}(p_i),\mathcal{T}_{\theta^{gt}}(p_i))<\tau\cdot\max(h,w)]}{\sum_{i=1}^n|p_i|},
	\end{equation}
	where $p_i$ is the $i${th} point, which consists of $(x_i,y_i)$, and~$\tau\cdot\max(h,w)$ refers to the tolerance term in the image size of $h\times w$. Intuitively, the~denominator and the numerator denote the number of correct keypoints and overall annotated keypoints, respectively. The~PCK metric shows how well matching is successful globally according to given $\tau$ with a lot of test images. In~this evaluation, we assess in the cases of $\tau=0.1, 0.3,$ and $0.5$. The~greater value of $\tau$ allows measuring degrees of matching more globally.
	To adopt the PCK metric, we annotated the keypoints and ground-truth transformation to 500 multi-temporal aerial image pairs. The~multi-temporal pairs are captured in Google Earth and composed of major administrative districts in South Korea, like the training image pairs.
	The annotation process is as the following process: (1) we extracted the keypoints of multi-temporal aerial image pairs using SIFT~\cite{Lowe04distinctiveimage}, and~(2) picked up the overlapping keypoints between each image pair. We annotate 20 keypoints per image pair, which generate a total of 10k keypoints for a quantitative assessment. This approach provides a fair demonstration of quantitative performance. In~the evaluation and the inference procedure, we used a two-stream network, except~for the augmented branch shown in Algorithms~\ref{alg:alg_2}.
	
	     \vspace{12pt}
    % Inference procedure
	\begin{algorithm}[H]
	    \label{alg:alg_2}
        \SetAlgoLined
        \SetKwInOut{Input}{Input}
        \Input{Source and target images $(I_S, I_T)$ \\ Trained model $\mathcal{M}_w$}
        \SetKwInOut{Output}{Output}
        \Output{Transformed image $I_{S}^{'}$}
        \texttt{\# Feed-forward}\\
        $\hat{\theta}_{S \rightarrow T}, \hat{\theta}_{T \rightarrow S}=\mathcal{M}_w(I_S, I_T)$\;
        \texttt{{\#} Ensemble}\\
        $\hat{\theta}_{en}=\mu(\hat{\theta}_{S \rightarrow T}, (\hat{\theta}_{T \rightarrow S})^{-1})$\;
        \texttt{{\#} Transform source image to target image}\\
        $I_{S}^{'}=\mathcal{T}_{\hat{\theta}_{en}}(I_S)$
           
         \caption{Inference~procedure.}
    \end{algorithm}

	\subsection{Results}
	\vspace{-6pt}
	\subsubsection{Quantitative~results}
	\label{sssec:quantitative}
	\subsubsection*{Aerial Image Dataset}
 Table~\ref{tab:tab_1} shows quantitative comparisons to the conventional computer vision methods (SURF~\cite{Bay_surf:speeded}, SIFT~\cite{Lowe04distinctiveimage}, ASIFT~\cite{Morel:2009:ANF:1658384.1658390} + RANSAC~\cite{Fischler:1981:RSC:358669.358692} and OA-Match~\cite{SONG2019317}) and CNNGeo~\cite{Rocco_2017_CVPR} on aerial image data with large transformation. Conventional computer vision methods~\cite{Bay_surf:speeded,Lowe04distinctiveimage,Morel:2009:ANF:1658384.1658390, Fischler:1981:RSC:358669.358692, SONG2019317} showed quite a number of critical failures globally. As~shown in Table~\ref{tab:tab_1}, the~conventional methods show low PCK performance in the case of $\tau=0.05$. However, in~the case of $\tau=0.01$, these methods showed lower degradation of performance compared with other deep learning based methods. This result implies that conventional methods enable finer matching if the matching procedure does not failed entirely. Although~CNNGeo fine-tuned by aerial images shows somewhat tolerable performance, our method considerably outperforms this method in all cases of $\tau$. Furthermore, we performed an investigation of the various backbone networks to demonstrate the importance of feature extraction. Since the backbone network substantially affects the total performance, we experimentally adopted the best backbone~network.
		
	% Comparisons of PCK on the aerial images
	\begin{table}[H]
    	\caption{Comparisons of probability of correct keypoints (PCK) in the aerial images. CNNGeo is evaluated in two versions: the pre-trained model provided in~\cite{Rocco_2017_CVPR} and the fine-tuned model by the aerial images. Both models use ResNet101 as the backbone~network.}
		\centering
			\begin{tabular}{lccc}
				\toprule

				\multirow{2}{*}{\textbf{Methods}} & \multicolumn{3}{c}{\textbf{PCK (\%)}} \\
				                         & \boldmath{$\tau=0.05$} 
				                         &  \boldmath{$\tau=0.03$ }
				                         &  \boldmath{$\tau=0.01$} \\
				\midrule
				SURF~\cite{Bay_surf:speeded} & 26.7 & 23.1 & 15.3 \\
				SIFT~\cite{Lowe04distinctiveimage} & 51.2 & 45.9 & 33.7 \\
				ASIFT~\cite{Morel:2009:ANF:1658384.1658390} & 64.8 & 57.9 & 37.9 \\
				OA-Match~\cite{SONG2019317} & 64.9 & 57.8 & 38.2 \\
				\midrule
				CNNGeo~\cite{Rocco_2017_CVPR}  (pretrained) & 17.8 & 10.7 & 2.5 \\
				CNNGeo (fine-tuned) & 90.6 & 76.2 & 27.6\\
				Ours; ResNet101~\cite{He_2016_CVPR} & {\bf 93.8} & {\bf 82.5} & {\bf 35.1}\\
				\midrule
				Ours; ResNeXt101 \cite {Xie_2017_CVPR}  & 94.6 & 85.9 & 43.2\\
				Ours; Densenet169~\cite{huang_2017_densely} & 95.6 & 88.4 & 44.0\\
				Ours; SE-ResNeXt101~\cite{Hu_2018_CVPR} &{\bf 97.1} & {\bf 91.1} & {\bf 48.0}\\
				\bottomrule

				\end{tabular}
		\label{tab:tab_1}
	\end{table}

\subsubsection*{Ablation Study}
	The proposed method combines two distinct techniques: (1) internal augmentation and (2) bidirectional ensemble. We analyze the contributions and effects of each proposed component and compare our models with CNNGeo~\cite{Rocco_2017_CVPR}. '+ Int. Aug.' and '+ Bi-En.’, which signify the internal augmentation and bidirectional ensemble addition, respectively. As~shown in Table~\ref{tab:tab_2}, all models added by our proposed component improves the performances of CNNGeo for all $\tau$, while maintaining the number of parameters.
	We further compare the proposed two-stream architecture to single-stream architecture which is added to the proposed components (internal augmentation, bidirectional ensemble). Table~\ref{tab:tab_3} shows the excellence of the proposed two-stream architecture compared to the single-stream architecture. It implies that the proposed regularization terms by the two-stream architecture are~reasonable.
	
	% Ablation study
	\begin{table}[H]
	\caption{{Results of models with different additional components.} We analyzed the contributions of each component with ResNet-101~backbone.}
	\centering
			\begin{tabular}{lccc}
				\toprule

				\multirow{2}{*}{\textbf{Methods}} & \multicolumn{3}{c}{\textbf{PCK~(\%)}} \\
				                         & \boldmath{$\tau=0.05$} &\boldmath{ $\tau=0.03$}&\boldmath{$\tau=0.01$} \\
				\midrule
				CNNGeo~\cite{Rocco_2017_CVPR} & 90.6 & 76.2 & 27.6 \\
				CNNGeo + Int. Aug. & 90.9 & 76.6 & 28.4\\
				CNNGeo + Bi-En. & 92.1 & 79.5 & 31.8\\
				CNNGeo + Int. Aug. + Bi-En. (Ours) &{\bf 93.8} & {\bf 82.5} & {\bf 35.1}\\
				\bottomrule

				\end{tabular}
		\label{tab:tab_2}
	\end{table}
\unskip
	
	\begin{table}[H]
	\caption{{Comparison of single-stream and two-stream architecture.} We analyzed the effectiveness of the two-stream based regularization with ResNet-101~backbone.}
		\centering
			\begin{tabular}{lccc}
				\toprule

				\multirow{2}{*}{\textbf{Methods}} & \multicolumn{3}{c}{\textbf{PCK~(\%)}} \\
				                         & \boldmath{$\tau=0.05$} & \boldmath{$\tau=0.03$} &\boldmath{$\tau=0.01$} \\
				\midrule
				Single-stream (with Int. Aug. and Bi-En.) & 92.4 & 79.7 & 33.5\\
				Two-stream (Ours) &{\bf 93.8} & {\bf 82.5} & {\bf 35.1}\\
					\bottomrule
				\end{tabular}
		\label{tab:tab_3}
	\end{table}
    
	\subsubsection{Qualitative~Results}
	\subsubsection*{Global Matching Performance}
	We performed a qualitative evaluation using the Google Earth dataset (Figure \ref{fig:fig_5}) and the ISPRS dataset (Figure \ref{fig:fig_6}). The~ISPRS dataset is a real-world aerial image dataset that was obtained from different viewpoints. Although~our model was trained from the synthetic transformed aerial image pairs, it is successful with real-world data. In~Figure~\ref{fig:fig_5} and \ref{fig:fig_6}, the~samples consist of challenging pairs, including numerous difficulties such as differences in time, occlusion, changes in vegetation, and~large-scale transformation between the source images and the target images. Our method correctly aligned the image pairs and yields accurate results of matching compared with other methods~\cite{Morel:2009:ANF:1658384.1658390, Fischler:1981:RSC:358669.358692, SONG2019317, Rocco_2017_CVPR} as shown in Figure~\ref{fig:fig_5} and \ref{fig:fig_6}. 
	
	% Qualitative results figure (GoogleEarth)
	\newcolumntype{C}[1]{>{\centering\arraybackslash}p{#1}}
	\begin{figure}[H]
	
       \centering
		\begin{tabular}{C{25mm}C{22mm}C{20mm}C{22mm}C{22mm}c}
            \multicolumn{6}{c}{\includegraphics[width=15.5cm]{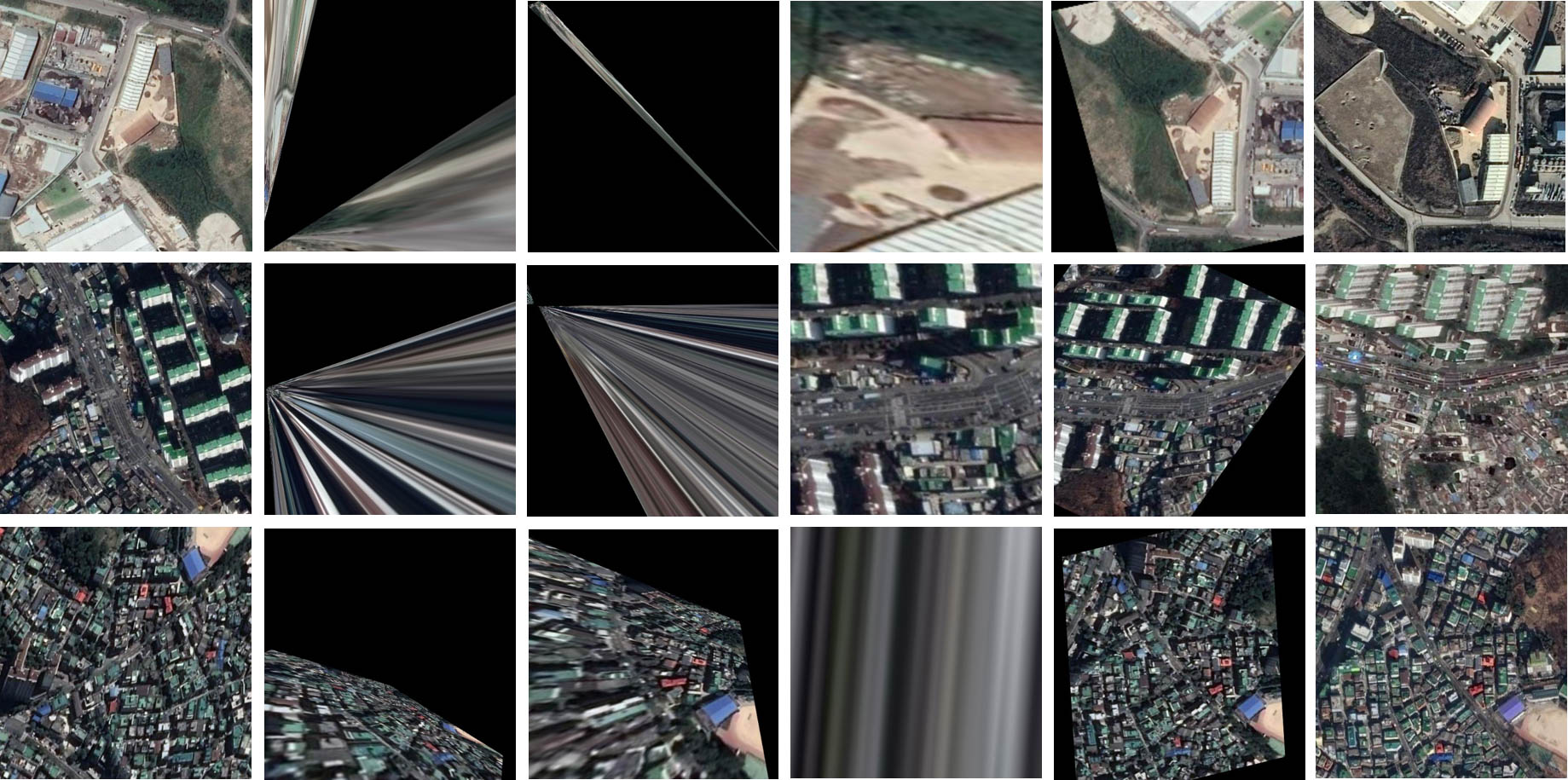}}
			\\
		\scriptsize{Source} & \scriptsize{ASIFT~\cite{Morel:2009:ANF:1658384.1658390} +}& \scriptsize{OA-Match~\cite{SONG2019317}} & \scriptsize{CNNGeo~\cite{Rocco_2017_CVPR}} & \scriptsize{Ours} & \scriptsize{Target} 
			\\
			& \scriptsize{RANSAC~\cite{Fischler:1981:RSC:358669.358692}}
		\end{tabular}
		\caption{{Qualitative results for Google Earth data.} These sample pairs are captured in Google Earth with different environments (viewpoints, times, and~sensors).}
		\label{fig:fig_5}
    \end{figure}
    
    % Qualitative results figure (ISPRS)
    \newcolumntype{C}[1]{>{\centering\arraybackslash}p{#1}}
	\begin{figure}[H]
       \centering
		\begin{tabular}{C{25mm}C{22mm}C{20mm}C{22mm}C{22mm}c}
            \multicolumn{6}{c}{\includegraphics[width=15.5cm]{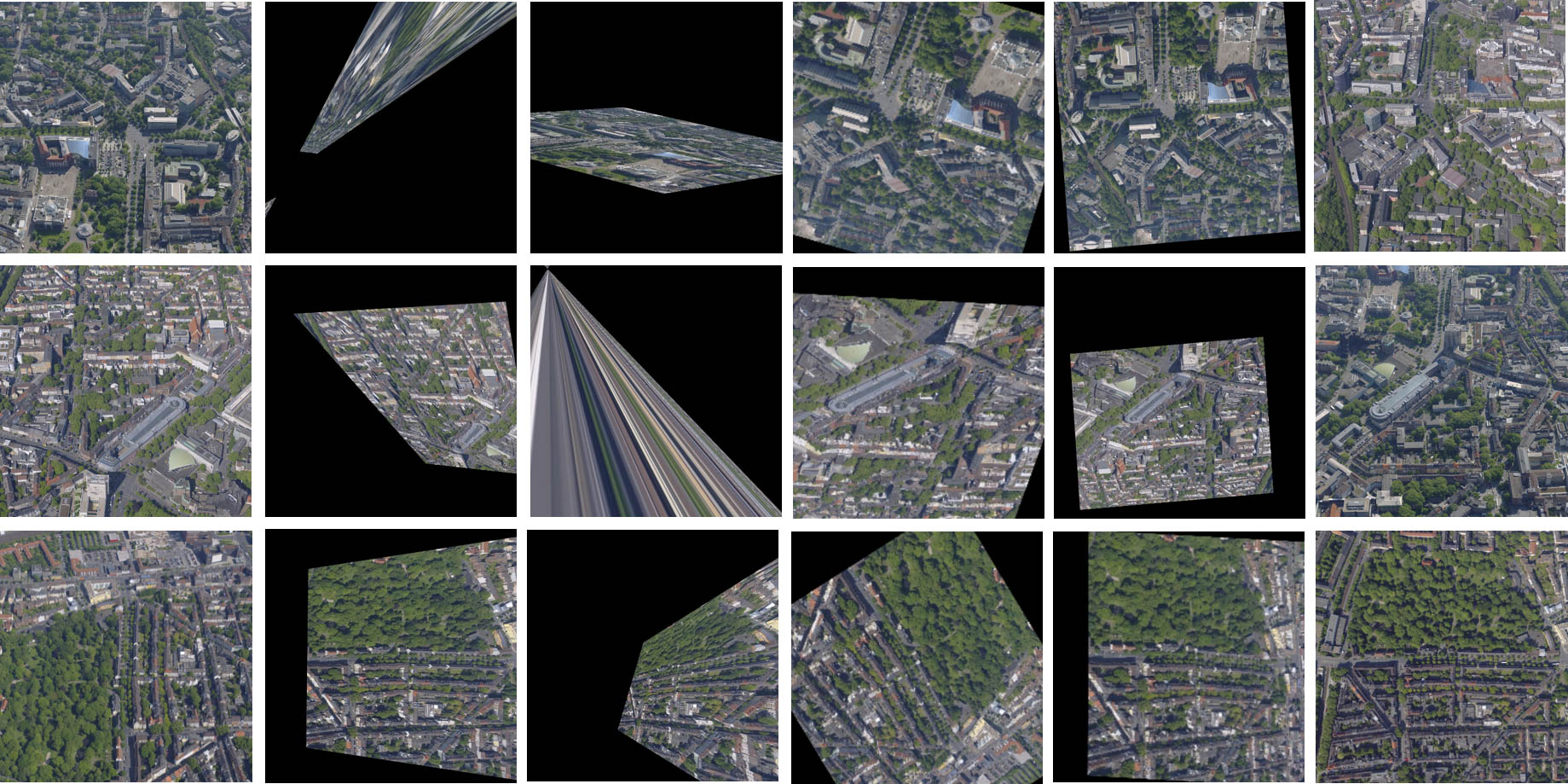}}
			\\
		\scriptsize{Source} & \scriptsize{ASIFT~\cite{Morel:2009:ANF:1658384.1658390} +}& \scriptsize{OA-Match~\cite{SONG2019317}} & \scriptsize{CNNGeo~\cite{Rocco_2017_CVPR}} & \scriptsize{Ours} & \scriptsize{Target} 
			\\
			& \scriptsize{RANSAC~\cite{Fischler:1981:RSC:358669.358692}}
		\end{tabular}
		\caption{{Qualitative results for the ISPRS dataset.} These samples are released by ISPRS~\cite{ISPRS}.}
		\label{fig:fig_6}
    \end{figure}
	
\subsubsection*{Localization Performance}
\begin{figure}[H]
	    \centering
	    \includegraphics[width=15.5cm]{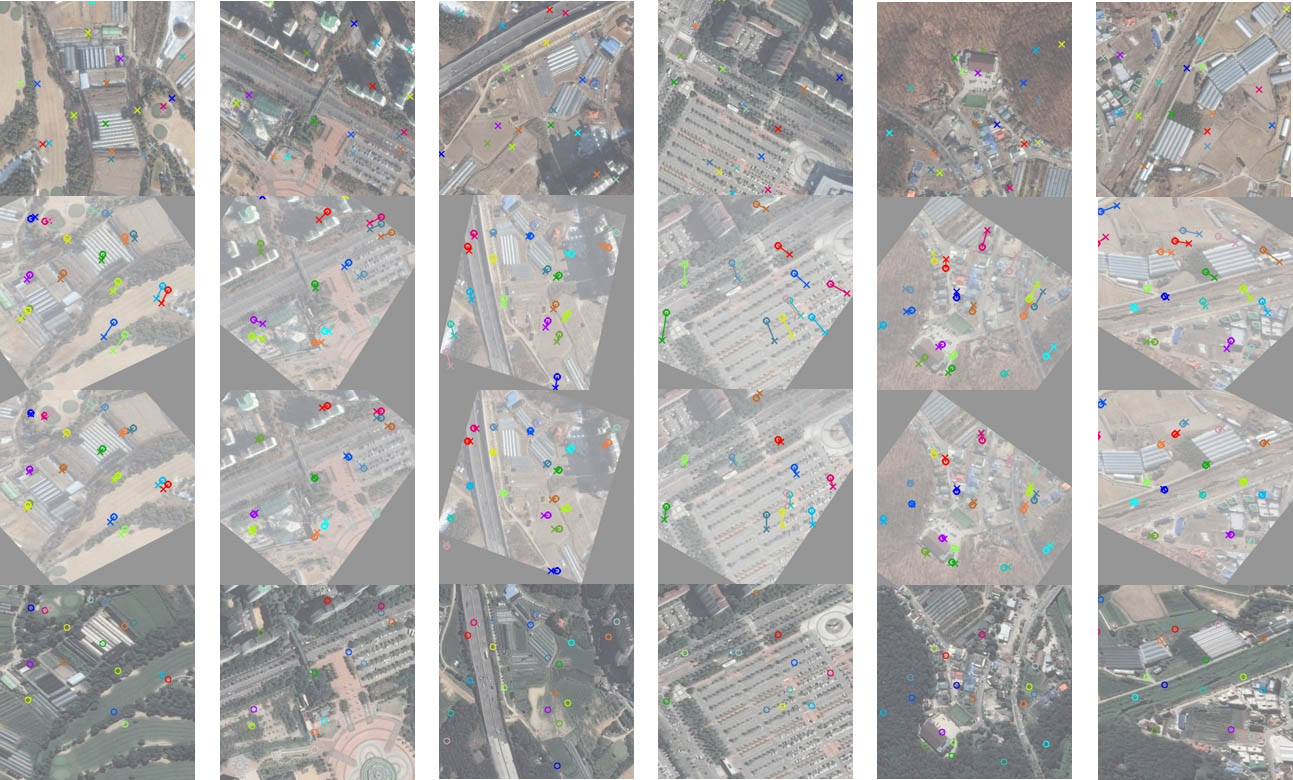}
	    \caption{{Visualization of the matched keypoints.} Rows are each as follows: (1) source images, (2) results of CNNGeo~\cite{Rocco_2017_CVPR}, (3) results of our method, (4) target~images.}
	    \label{fig:visualize_keypoints}
	\end{figure}

	We visualized the matched keypoints for comparing localization performance with CNNGeo~\cite{Rocco_2017_CVPR}. It is also important how fine source and target images are matched within the success cases. As~shown in Figure~\ref{fig:visualize_keypoints}, we intuitively compared localization performance. The~X marks and the O marks on the images indicate the keypoints of the source images and the target images, respectively. Both models (\cite{Rocco_2017_CVPR} and ours) successfully estimated global transformation. However, looking at the distance of matched keypoints, ours was better~localized.
	
	% Visualization of matched keypoints
\unskip

\section{Discussion}
\unskip
	\subsection{Robustness for the Variance of Aerial~Image}
\begin{figure}[H]
	    \centering
	    \includegraphics[width=13.5cm]{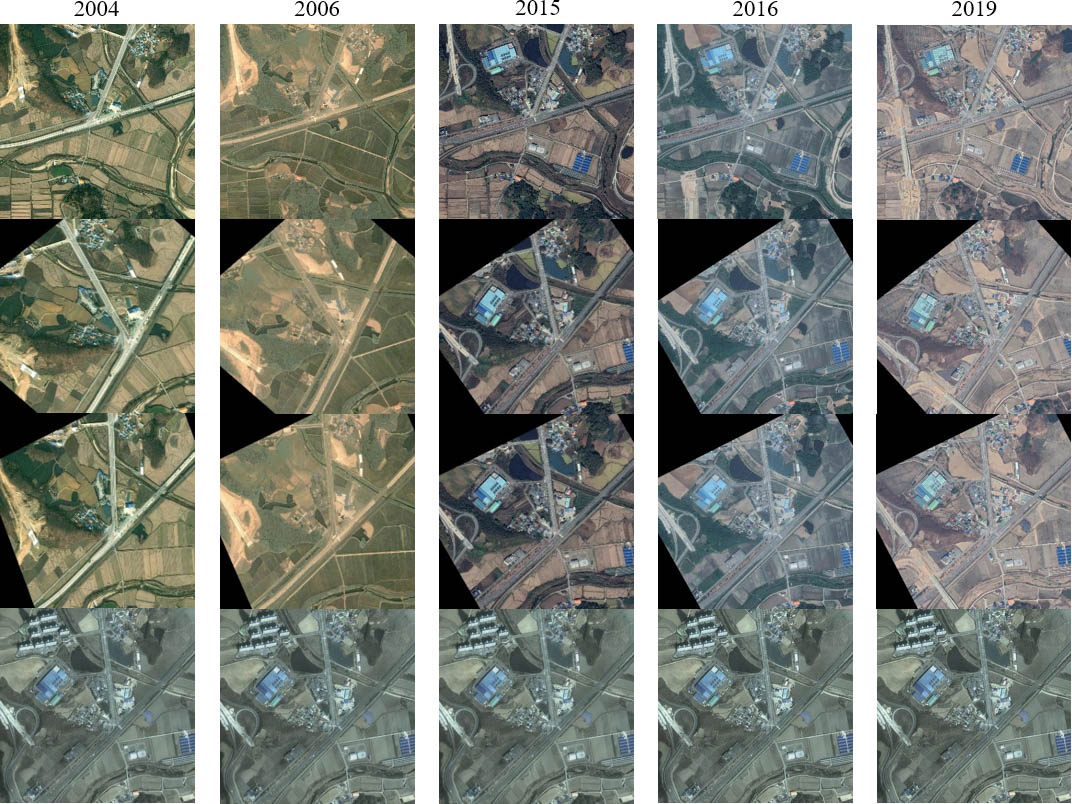}
	    \caption{{Results for various source images taken at different times.} Rows are each as follows: (1) source images, (2) results of CNNGeo~\cite{Rocco_2017_CVPR}, (3) results of our method, (4) target~images.}
	    \label{fig:robustness}
	\end{figure}

	Furthermore, we experimented on robustness for the variance of aerial images as shown in Figure~\ref{fig:robustness}. The~source images were taken in 2004, 2006, 2015, 2016, and~2019, respectively. The~target images were absolutely identical images. As~a result, ours showed more stable results for overall sessions. Especially, source images which were taken in 2004 and 2006 have large differences of including object compared with the target image. It showed that ours had better robustness for the variance of the aerial images while the baseline~\cite{Rocco_2017_CVPR} is significantly influenced by these~differences. 
	\subsection{Limitations and Analysis of Failure~Cases}
	\label{sssec:limitation}
	We describe the limitation of our method and analyze the case in which the proposed method fails. As~shown in Section~\ref{sssec:quantitative}, our method quantitatively showed state-of-the-art performance. However, comparing $\tau=0.05$ with $\tau=0.01$ indicates a substantial difference in performance. Our method is weak in detailed matching even though it successfully estimates global transformation in most cases. This weakness can be addressed by additional fine-grained transformation as~post-processing. 
	
\begin{figure}[H]
       \centering
		\includegraphics[width=11.5cm]{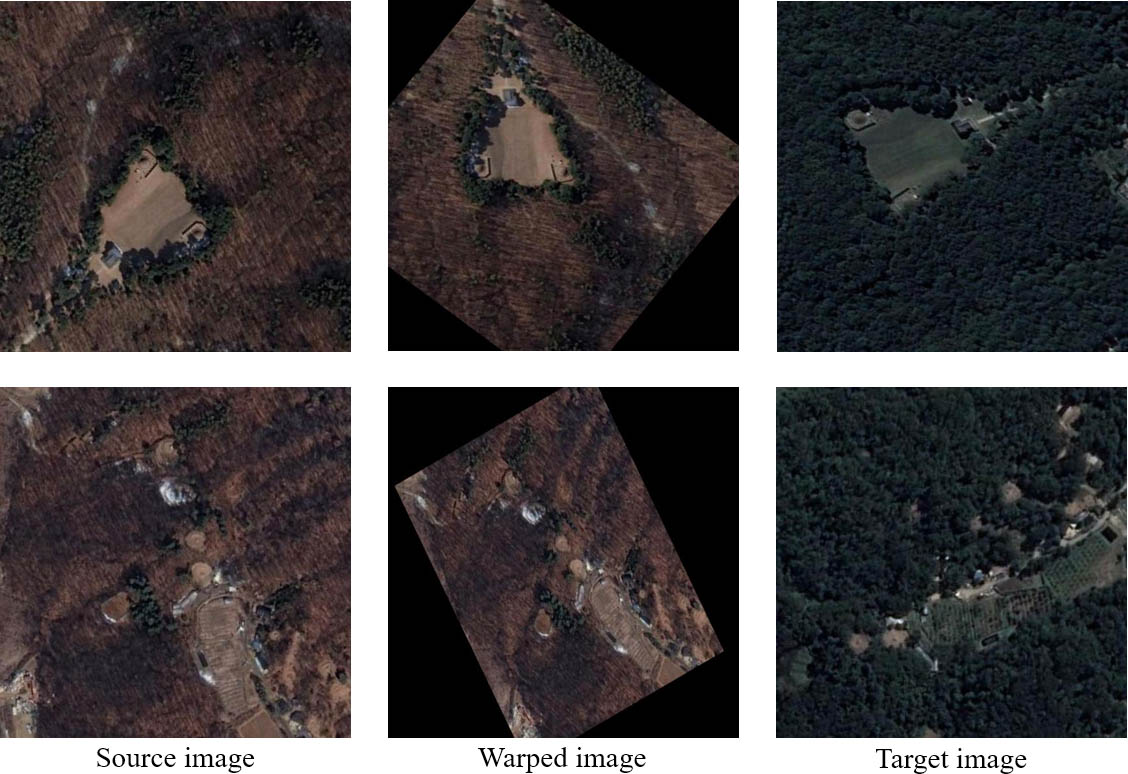}
		\caption{{Failure cases, which primarily consist of wooded areas.} Although there are objects that can be focused, it fails completely. \vspace{0.5cm}}
		\label{fig:fail_wood}
    \end{figure}

\begin{figure}[H]
       \centering
		\includegraphics[width=11.5cm]{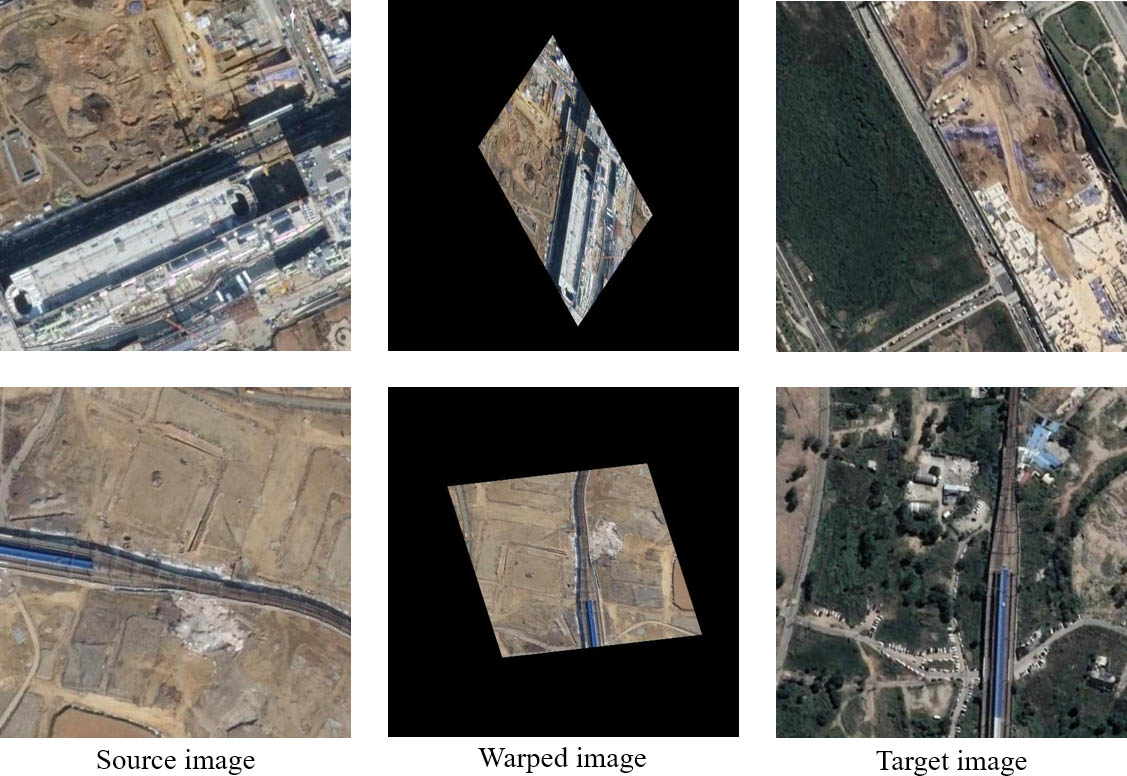}
		\caption{{Failure cases, which are largely changed areas.} Since the changed area is too large, it fails~completely.}
		\label{fig:fail_large_changed}
    \end{figure}

	Our proposed method failed in several cases. As~a result, we have determined that our method fails in mostly wooded areas or largely changed areas as shown in Figure~\ref{fig:fail_wood} and \ref{fig:fail_large_changed}. In~mostly wooded areas, repetitive patterns hinder the focus on a salient region. In~the case of largely changed areas, massive differences, such as buildings, vegetation, and~land-coverage between the source image and the target image are observed, which leads to degradation of performance. To~address these limitations, a~method that can aggregate local contexts for reducing repetitive patterns is~required. 
	% Robustness for the variance of aerial image
\unskip

	% Failure cases figure (wooded area)
    
    % Failure cases figure (largely changed area)
\unskip

%%%%%%%%% Conclusion
\section{Conclusions}
    \label{sec:conclusion}
    \textls[-5]{We propose a novel approach based on a deep end-to-end network for aerial image matching. To~become robust to the variance of the aerial images, we introduce two-stream architecture using internal augmentation. We show its efficacy for consideration of various image pairs. An~augmented image can be seen as an image which is taken in different environments (brightness, contrast, saturation), and~by training these images with original target images simultaneously, it leads to the effect of regularizing the deep network. Furthermore, by~training and inferring in two possible directions, we apply an efficient ensemble method without any additional networks or parameters, which considers the variances between transformation parameters from both directions and substantially improves performance. In~the experimental section, we show stable matching results with a large volume of aerial images. However, our method also has some limitations as aforementioned (Section~\ref{sssec:limitation}). To~overcome these limitations, we plan to research the localization problem and the attention mechanism. Moreover, The studies applying Structure from Motion (SfM) and 3D reconstruction to image matching are very interesting and can improve performance of image matching, so we also plan to conduct this study in the future work.}

%%%%%%%%%%%%%%%%%%%%%%%%%%%%%%%%%%%%%%%%%%
\vspace{6pt}
\authorcontributions{conceptualization, J.-H.P., W.-J.N. and S.-W.L.; data curation, J.-H.P. and  W.-J.N.; formal analysis, J.-H.P. and  W.-J.N.;
funding acquisition, S.-W.L.; investigation, J.-H.P. and  W.-J.N.; methodology, J.-H.P. and  W.-J.N.; project administration, S.-W.L.; resources, S.-W.L.;
software, J.-H.P. and  W.-J.N.; supervision, S.-W.L.; validation, J.-H.P.,  W.-J.N. and S.-W.L.; visualization, J.-H.P. and  W.-J.N.; writing---original draft, J.-H.P. and  W.-J.N.; writing---review and editing, J.-H.P.,  W.-J.N. and S.-W.L. All authors have read and agreed to the published version of the manuscript.}

\funding{This work was supported by the Agency for Defense Development	(ADD) and the Defense Acquisition Program Administration (DAPA) of Korea (UC160016FD).}

%%%%%%%%%%%%%%%%%%%%%%%%%%%%%%%%%%%%%%%%%%
\acknowledgments{The authors would like to thank the anonymous reviewers for their valuable suggestions to improve the quality of this paper.}

%%%%%%%%%%%%%%%%%%%%%%%%%%%%%%%%%%%%%%%%%%
\conflictsofinterest{The authors declare no conflicts of~interest.} 
\newpage

%%%%%%%%%%%%%%%%%%%%%%%%%%%%%%%%%%%%%%%%%%%
\abbreviations{The following abbreviations are used in this manuscript:\\

\noindent 
\begin{tabular}{@{}ll}
DNNs & Deep Neural Networks\\
CNNs & Convolutional Neural Networks\\
ReLU & Rectified Linear Unit\\
TPS & Thin-Plate Spline\\
PCK & Probability of Correct Keypoints\\
ADAM & ADAptive Moment estimation\\
Bi-En. & Bidirectional Ensemble\\
Int. Aug. & Internal Augmentation\\
ISPRS &  International Society for Photogrammetry and Remote Sensing
\end{tabular}}

%%%%%%% References
\reftitle{References}
\bibliography{egbib}
\end{document}